%% file: LeFT-nips.tex
\documentclass{article}

\PassOptionsToPackage{numbers,compress}{natbib}
\usepackage[preprint]{neurips_2025}   
\makeatletter
\renewcommand{\@notice}{}
\makeatother

\usepackage[utf8]{inputenc}
\usepackage[T1]{fontenc}
\usepackage{amsmath,amssymb,amsfonts}
\usepackage{nicefrac}
\usepackage{xcolor}
\usepackage{booktabs}
\usepackage{graphicx}
\usepackage{array}
\usepackage{caption}
\usepackage{enumitem}
\usepackage{url}
\usepackage[hidelinks]{hyperref}
\usepackage{microtype}
\usepackage{placeins}

\hypersetup{
    pdftitle={Learn, Freeze, and Transfer an Activation Function Across Architectures},
    pdfauthor={Arun Prakash A},
    pdfsubject={Transfer of Globally Shared Learned Activation Functions},
    pdfkeywords={Learnable Activations, Transfer Learning,
                 Fourier Activation, Vision Transformers}
}

\graphicspath{{figures/}}

\newcommand{\fact}{FAct}
\newcommand{\ffact}{Frozen \fact{}}
\newcommand{\efact}{Learnable \fact{}}

\newcommand{\pp}{\ensuremath{\,\mathrm{pp}}}
\newcommand{\vitsix}{ViT-d6}
\newcommand{\vittwo}{ViT-d2}

\title{Learn, Freeze, and Transfer an Activation Function\\Across Architectures}

\author{%
  Arun Prakash A\\
  Nilekani Centre at AI4Bharat\\
  Indian Institute of Technology Madras\\
  \texttt{a.arun283@gmail.com}%
}

\begin{document}
\maketitle
\begingroup\renewcommand\thefootnote{}
\footnotetext{%
\setlength{\tabcolsep}{0pt}%
\begin{tabular}{@{}>{\raggedright\arraybackslash}p{0.45\linewidth}@{\hspace{0.06\linewidth}}>{\raggedright\arraybackslash}p{0.45\linewidth}@{}}
Code & Paper with interactive figures\\
\url{https://github.com/AI4Bharat/LeFT-activation-transfer} &
\url{https://arunprakash-a.github.io/papers/learn-freeze-transfer/}
\end{tabular}}
\endgroup

\begin{abstract}
Transfer learning typically reuses knowledge acquired by a source network through its learned weights.
We investigate whether a learned activation function can also be transferred and remain useful in
networks trained from scratch. We study this complementary form of transfer through a learn--freeze--transfer
protocol, in which the activation is optimized jointly with a source network in a single
training run, and is then frozen and reused in target networks trained from scratch. A pointwise
activation with globally shared parameters is independent of layer dimensions and can therefore be
incorporated into networks of different architectures trained from scratch. We instantiate this protocol
with \fact{}, a five-parameter activation represented by a two-harmonic truncated Fourier series,
initialized from a Fourier approximation of GELU and learned in a Vision Transformer (ViT) on
ImageNet-1K. When reused in a ViT of the same architecture trained from scratch on ImageNet-1K, the
frozen activation (\ffact{}) improves top-1 accuracy by 1.46 percentage points over GELU. Across four
smaller target architectures (standard ViT, Swin Transformer, linear-attention ViT, and MLP-Mixer) and
four image-classification datasets, it achieves the highest mean accuracy among eleven activation
functions in 15 of 16 settings under a common training configuration, with improvements of up to 1.71
percentage points over the best conventional activation. When the learning rate and weight decay are
tuned separately for each activation under the same search budget, \ffact{} still ranks first in 11 of
the 16 settings, second in four, and fourth in one. These results provide
empirical evidence that a learned activation function can retain its usefulness when transferred across vision
architectures, independently of the weights with which it was learned.
\end{abstract}

\section{Introduction}
\label{sec:intro}

Transfer learning typically reuses knowledge acquired by a source network through its learned weights.
We investigate whether an activation function learned jointly with a source network in a single training
run can also be transferred to target networks and remain useful. To this end, we
use a parameterized activation function whose coefficients are shared globally across all activation
sites of the network. Global sharing makes the number of activation parameters independent of network
depth and width, so the learned function can be reused in networks of different architectures. This
yields a complementary form of transfer, in which the learned activation, rather than the source-network
weights, is retained and reused.

Learnable activation functions allow the nonlinear transformation to adapt during training. Existing
parameterizations include adaptive slopes~\citep{he2015prelu,jagtap2020gaaf}, customized activation
functions~\citep{ma2021acon}, rational functions~\citep{molina2020pau}, and kernel-based
representations~\citep{scardapane2019kaf}. Their parameters may be assigned locally to individual
neurons or layers~\citep{agostinelli2015apl,he2015prelu,jagtap2020laaf,ma2021acon,molina2020pau} or
shared across multiple layers~\citep{jagtap2020gaaf,delfosse2024rational}. These approaches study how
activation functions adapt within a network. In contrast, we examine whether a learned activation can be
retained and reused independently of the network in which it was optimized.

\begin{figure}[t]
\centering
\includegraphics[width=\linewidth]{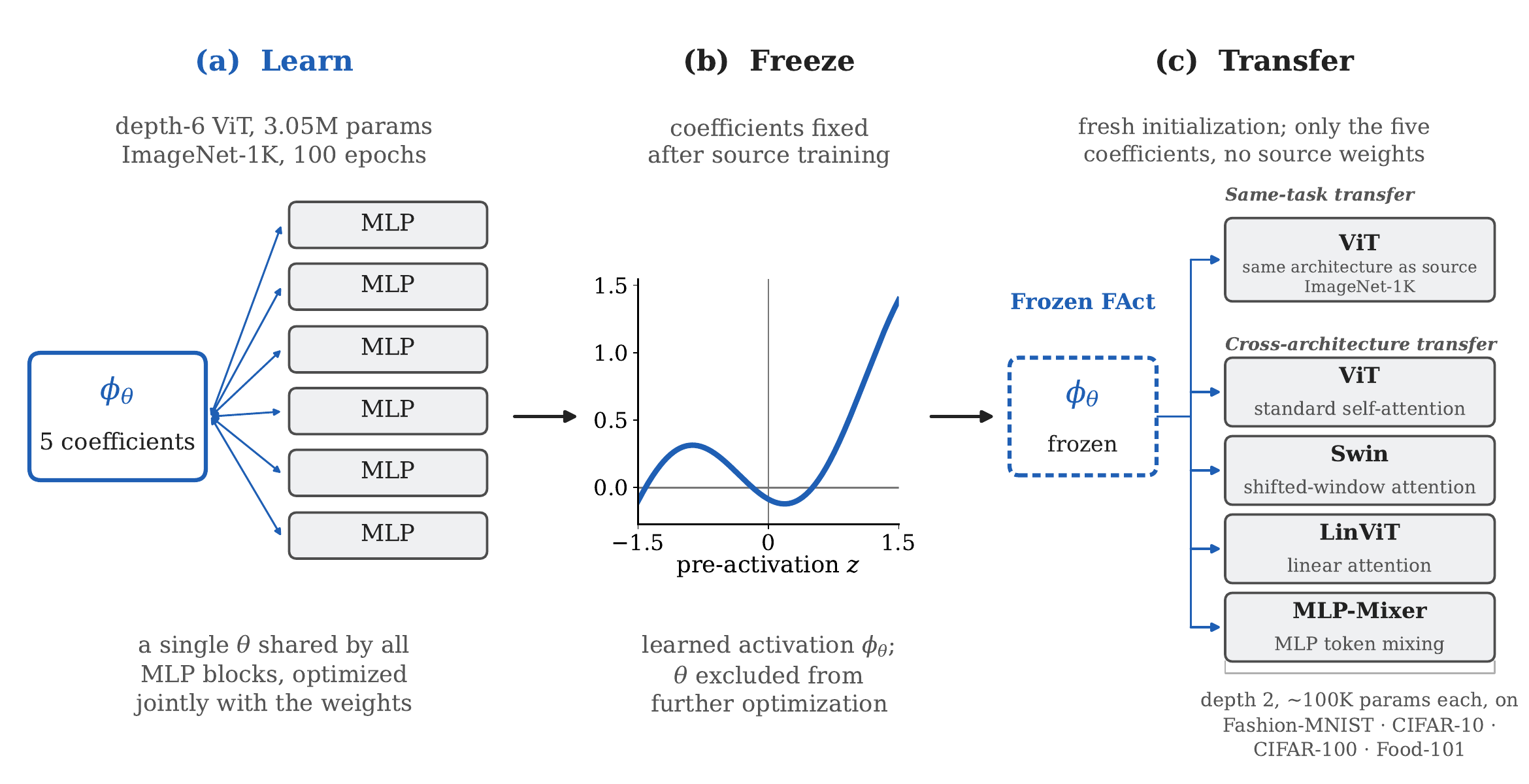}
\caption{\textbf{Overview of the learn--freeze--transfer framework.} (a) A five-parameter \fact{}
activation, shared across all MLP blocks, is learned jointly with a depth-6 ViT on ImageNet-1K.
(b) The learned activation coefficients are frozen after source training. (c) Only these
coefficients, not the source-network weights, are transferred to independently initialized target
networks. Evaluation includes same-task transfer to a freshly initialized copy of the source
architecture (\vitsix{}) and cross-architecture transfer to depth-2 targets (\vittwo{}, Swin-d2,
LinViT-d2, and MLP-Mixer), with the activation coefficients fixed throughout target training.
Transfer performance for these settings is reported in
Tables~\ref{tab:same-task}, \ref{tab:transfer-shared}, \ref{tab:swin-main}, \ref{tab:linvit-main},
and~\ref{tab:mixer}.}
\label{fig:pipeline}
\end{figure}

Prior work has reused activation functions beyond the setting in which they were obtained, but mainly
through automated search~\citep{ramachandran2018swish,bingham2022pangaea,strack2024search} or
meta-learning~\citep{teney2025simplicity}. To our knowledge, transferring an activation learned in a
single training run across architectures has not been studied. In our setting, the activation is learned
through end-to-end training with a single source network, without meta-learning or a separate search.
After source training, the activation is frozen and transferred to target networks trained from scratch.
Only its learned coefficients are transferred; the source-network weights are not reused, and the
activation remains fixed throughout target training (Figure~\ref{fig:pipeline}). This protocol allows us
to examine whether a nonlinearity learned within one network remains useful across different
architectures and tasks.

We instantiate this protocol with \fact{} (Fourier Activation), a learnable activation represented by a
two-harmonic truncated Fourier series with five coefficients. A single set of coefficients is shared
across all MLP blocks of the network. The coefficients are initialized from a two-harmonic Fourier
approximation of GELU~\citep{hendrycks2016gelu} and optimized jointly with the source-network weights.
We refer to the activation after source training, with its coefficients fixed, as \ffact{}.

We evaluate \ffact{} in two transfer settings. In the same-task setting, the activation learned by a
depth-6 Vision Transformer (ViT)~\citep{dosovitskiy2021vit} on ImageNet-1K~\citep{deng2009imagenet} is
reused in a ViT of the same architecture, trained from scratch on the same task. In the
cross-architecture setting, the same five coefficients are used in four depth-2 target architectures: a
ViT with global self-attention, a Swin Transformer with shifted-window self-attention~\citep{liu2021swin},
a ViT with linear attention~\citep{katharopoulos2020linear}, and an attention-free
MLP-Mixer~\citep{tolstikhin2021mixer}. These targets contain approximately 100K parameters each and are
trained on four image-classification datasets: Fashion-MNIST~\citep{xiao2017fmnist}, CIFAR-10,
CIFAR-100~\citep{krizhevsky2009cifar}, and Food-101~\citep{bossard2014food101}.

In the same-task setting, \ffact{} improves ImageNet-1K top-1 accuracy by 1.46 percentage points over
GELU. In the cross-architecture setting, it achieves the highest mean test accuracy among eleven
activation functions in 15 of the 16 architecture--dataset combinations under a common training
configuration, with improvements of up to 1.71 percentage points over the best conventional activation.
When the learning rate and weight decay are tuned separately for each activation under the same search
budget, \ffact{} still ranks first in eleven combinations, second in four, and fourth in one. Finally, on
Swin targets of one to six blocks, a frozen, untrained Fourier approximation of GELU does not reproduce
these gains, indicating that the benefit comes from the learned activation rather than the Fourier
parameterization alone. In contrast to the SPLASH ablation of \citet{tavakoli2021splash}, these results
show that a frozen learned activation can retain its benefit after transfer.

This work makes three contributions:

\begin{enumerate}[leftmargin=1.5em,itemsep=3pt,topsep=3pt]
\item \textbf{A learn--freeze--transfer protocol for activation functions.} We formulate a complementary
form of transfer in which a globally shared activation is learned jointly with a source network in a
single training run, then frozen and reused in target networks trained from scratch, without
transferring source-network weights.

\item \textbf{A globally shared activation suited for transfer.} We instantiate the protocol with
\fact{}, whose five coefficients are shared across all MLP blocks, so that the learned activation can be
placed in networks of different architectures.

\item \textbf{An empirical evaluation of activation transferability.} We evaluate same-task and
cross-architecture transfer using three transformer architectures and an attention-free MLP-Mixer on
four image-classification datasets. We compare \ffact{} with ten conventional activation functions under
common and activation-specific hyperparameter configurations, and we include a control showing that the
gains come from the learned activation rather than its parameterization.
\end{enumerate}

\section{Related Work}
\label{sec:related}

\subsection{Transfer Learning}
\label{sec:transfer-learning}

Transfer learning reuses knowledge acquired on a source task to improve learning on a target
task~\citep{pan2010survey}. In deep networks, this is typically done by reusing learned weights, either
to initialize a target network that is then fine-tuned or as fixed feature
extractors~\citep{yosinski2014transferable}. Knowledge distillation transfers knowledge between networks
of different architectures, but it requires the source network during target training to provide
supervision~\citep{hinton2015distilling}. In both cases, the transferred knowledge is carried by the
source network's weights or outputs. We study a complementary setting in which only a learned activation
function is transferred, and the source network is not used during target training.

\subsection{Learnable and Globally Shared Activation Functions}
\label{sec:learnable}

Learnable activation functions introduce parameterized nonlinear transformations whose parameters are
optimized jointly with the network weights. PReLU learns the slope of the negative input
region~\citep{he2015prelu}, and ACON provides a more flexible parametric
formulation~\citep{ma2021acon}. Pad\'e activation units represent the nonlinearity as a ratio of
polynomials~\citep{molina2020pau}, while other approaches use adaptive piecewise-linear
functions~\citep{agostinelli2015apl} or kernel-based representations~\citep{scardapane2019kaf}.
Learnable Fourier-type activations have also been studied, with coefficients learned within the network
in which they are used~\citep{khalfaoui2026poly,ko2026tesla,shor2026spectral}.

Activation parameters also differ in how they are shared. They may be assigned to individual neurons or
layers, or shared across multiple layers. \citet{jagtap2020gaaf} introduced globally adaptive activation
functions for deep and physics-informed neural networks, using a single trainable scaling parameter that
adjusts the activation slope across the network. \citet{delfosse2024rational} shared a common set of
rational-function parameters across layers to regulate activation plasticity in deep reinforcement
learning. These approaches show that sharing activation parameters within a network can be useful.
However, sharing alone does not show whether a learned activation can be separated from its source
network and reused in another network.

\subsection{Reuse and Transfer of Activation Functions}
\label{sec:transfer-related}

Several studies have reused activation functions beyond the setting in which they were obtained.
\citet{ramachandran2018swish} discovered Swish by automated search on a small network and showed that it
transfers to larger models and other tasks. \citet{bingham2022pangaea} used surrogate-based search to
find activation functions that transfer from small search tasks to larger models, including ResNet-50 on
ImageNet. \citet{strack2024search} used gradient-based search and transferred the resulting activations,
with their constants fixed, to larger models of the same type and to new datasets.
\citet{teney2025simplicity} meta-learned activation functions through bilevel optimization across task
episodes, demonstrating transfer across algorithmic tasks and reuse in networks of different widths.

Reusing a learned shape does not always preserve its benefit. \citet{tavakoli2021splash} found that a
fixed activation mimicking the final shape of trained SPLASH units performed only on par with ReLU, and
concluded that the evolution of the shape during training was important. Although this was not a transfer
study, it suggests that freezing a learned activation may discard part of its benefit.

The present study differs from these works in two respects. First, the activation is learned in a single
conventional training run, without search or meta-learning. Second, the frozen coefficients are
transferred across architecture families, from a ViT to Swin, linear-attention, and MLP-Mixer targets,
without reusing source-network weights or updating the activation during target training.

\section{\fact{}: Learning and Transferring a Shared Nonlinearity}
\label{sec:method}

\subsection{Global Activation Sharing}
\label{sec:globalsharing}

Let $z_\ell$ denote the pre-activation at the $\ell$-th activation site in a network, and let
$\phi(z;\theta)$ denote a parameterized activation function. In \fact{}, a single parameter vector
$\theta$ is shared across all MLP blocks of the network:
\begin{equation}
h_\ell = \phi(z_\ell;\theta), \qquad \ell=1,\ldots,L,
\label{eq:sharing}
\end{equation}
where $L$ denotes the number of activation sites and $h_\ell$ is the corresponding activation output.
The same parameter vector is used at every site, without replication across layers or hidden units.

Because the activation parameters are shared, their gradients aggregate contributions from all
activation sites during training:
\begin{equation}
\frac{\partial \mathcal{J}}{\partial \theta}
=
\sum_{\ell=1}^{L}
\frac{\partial \mathcal{J}}{\partial h_\ell}\,
\frac{\partial \phi(z_\ell;\theta)}{\partial \theta},
\label{eq:gradient}
\end{equation}
where $\mathcal{J}$ denotes the network training objective. Each term represents the gradient
contribution propagated through an individual activation site. Consequently, the shared activation
parameters are optimized using signals from all MLP blocks rather than through independently learned,
layer-specific parameter sets.

Global sharing defines a single parameterized nonlinear function that can be retained independently
of the source-network weights. We refer to \fact{} while its parameters are jointly optimized with the
source-network weights as \textbf{\efact{}}. After source training, the parameters are frozen, and the
resulting activation is referred to as \textbf{\ffact{}}.

\subsection{Fourier Parameterization and Initialization}
\label{sec:fourier}

We parameterize \fact{} using a truncated Fourier series:
\begin{equation}
\phi(z;\theta)=a_0+\sum_{k=1}^{K}
\left[a_k\cos(k\omega z)+b_k\sin(k\omega z)\right],
\label{eq:fact}
\end{equation}
where $a_0$, $a_k$, and $b_k$ are learnable coefficients, and $K$ denotes the number of harmonics. We
fix the angular frequency to $\omega=1$, giving the activation a period of $2\pi$. The parameter
vector is defined as
\begin{equation}
\theta=(a_0,a_1,\ldots,a_K,b_1,\ldots,b_K),
\label{eq:theta}
\end{equation}
containing $2K+1$ coefficients. In the main experiments, we set $K=2$, resulting in five learnable
parameters shared across the entire network. Consequently, the number of activation parameters is
independent of network depth and width.

The Fourier representation provides a compact and smooth parameterization whose shape can adapt
during training. Depending on the learned coefficients, the resulting function can exhibit monotonic
or non-monotonic behaviour over the input range of interest. Furthermore, its output is bounded for
any finite coefficient values:
\begin{equation}
|\phi(z;\theta)|\leq |a_0|+\sum_{k=1}^{K}\bigl(|a_k|+|b_k|\bigr).
\label{eq:bound}
\end{equation}
This bound follows from the bounded magnitudes of the sine and cosine functions.

We initialize \fact{} using a two-harmonic Fourier approximation of GELU~\citep{hendrycks2016gelu}.
For a reference activation function $\sigma$, the Fourier coefficients are obtained by projection over
one period, $[-\pi,\pi]$:
\begin{subequations}
\label{eq:fourierfit}
\begin{align}
a_0&=\frac{1}{2\pi}\int_{-\pi}^{\pi}\sigma(z)\,dz, \label{eq:fourierfit-a0}\\
a_k&=\frac{1}{\pi}\int_{-\pi}^{\pi}\sigma(z)\cos(k\omega z)\,dz, \quad k=1,\ldots,K,
\label{eq:fourierfit-ak}\\
b_k&=\frac{1}{\pi}\int_{-\pi}^{\pi}\sigma(z)\sin(k\omega z)\,dz, \quad k=1,\ldots,K.
\label{eq:fourierfit-bk}
\end{align}
\end{subequations}
For $\sigma=\mathrm{GELU}$, $K=2$, and $\omega=1$, the resulting initialization is
\begin{equation}
\theta_0=(0.7061,\,{-0.7049},\,0.0261,\,1.0000,\,{-0.5000}).
\label{eq:init}
\end{equation}
This initialization provides a reference nonlinear transformation while allowing all five
coefficients to adapt during source-network training.

\subsection{Learn--Freeze--Transfer Protocol}
\label{sec:freeze}

The globally shared parameterization allows the learned activation to be separated from the source
network and reused independently. During source training, the source-network weights and shared
activation parameters are jointly optimized:
\begin{equation}
(W_s^*,\theta^*)
=
\arg\min_{W_s,\theta}
\mathcal{J}_{\mathrm{src}}(W_s,\theta),
\label{eq:sourceopt}
\end{equation}
where $W_s$ denotes the source-network weights, $\theta$ represents the shared activation parameters,
and $\mathcal{J}_{\mathrm{src}}$ is the source-task training objective. The resulting activation is
\begin{equation}
\phi^*(z)=\phi(z;\theta^*).
\label{eq:frozen}
\end{equation}

After source training, the learned activation parameters $\theta^*$ are frozen. Only these
parameters are transferred to a target network; the source-network weights $W_s^*$ are not
transferred. Let $W_t$ denote the weights of an independently initialized target network, and let
$\mathcal{J}_{\mathrm{tgt}}$ denote its training objective. Target training then optimizes
\begin{equation}
W_t^*
=
\arg\min_{W_t}
\mathcal{J}_{\mathrm{tgt}}\bigl(W_t;\theta^*\bigr),
\qquad
\theta^* \text{ fixed}.
\label{eq:targetopt}
\end{equation}
Thus, the target-network weights are optimized for the target task while the transferred activation
parameters remain unchanged throughout training. No source-network weights are used to initialize
the target model, and target-task gradients do not update the transferred activation.

This protocol isolates the reuse of the learned nonlinearity from the transfer of network weights.
It therefore enables us to examine whether an activation learned jointly with one source network
remains useful when incorporated into independently initialized networks, without further adaptation
of the activation itself. Figure~\ref{fig:pipeline} illustrates the three stages of this process.

\section{Experimental Evaluation}
\label{sec:experiments}

This section evaluates whether a globally shared activation learned jointly with a source network
remains useful when transferred to independently initialized target networks. We first describe the
source and target architectures, datasets, and training protocol. We then examine the evolution of the
source-learned activation, its transfer to a network with the same architecture and task, and its
transfer across four target architectures with different attention and token-mixing mechanisms.
\ffact{} is compared with ten conventional activation functions to assess its performance under a
common training configuration.

\subsection{Experimental Setup}
\label{sec:setup}

\paragraph{Source and target networks.}
We learn \fact{} in a depth-6 Vision Transformer (\vitsix{}; throughout, the suffix \mbox{-d$k$}
denotes a network of depth~$k$) trained on ImageNet-1K~\citep{deng2009imagenet}. The source network contains $3{,}048{,}232$ parameters, and
learning \fact{} adds five trainable activation coefficients, resulting in $3{,}048{,}237$ trainable
parameters. The same five coefficients are shared across all MLP blocks of the source network.
We train the source network with five random seeds (seeds~1--5); each run jointly optimizes the
source-network weights and the shared activation from the same GELU-based initialization. Unless
otherwise stated, the transfer experiments reuse the activation learned by the seed-1 source
checkpoint; the five source runs converge to nearly identical activations, as quantified in
Section~\ref{sec:evolution}, so this choice is representative rather than specific to one source run.

After source training, the learned activation is transferred to four independently initialized target
architectures: a depth-2 Vision Transformer (\vittwo{}) with global self-attention, a depth-2 Swin
Transformer (Swin-d2) with shifted-window self-attention~\citep{liu2021swin}, a depth-2 Vision
Transformer with linear attention (LinViT-d2)~\citep{katharopoulos2020linear}, and a depth-2
MLP-Mixer~\citep{tolstikhin2021mixer}. Each target network contains approximately 100K parameters, with
exact configurations provided in Appendix~\ref{app:protocol}. No source-network weights are
transferred, and the activation coefficients remain fixed throughout target training.

\paragraph{Datasets and transferred activation.}
Each target architecture is evaluated on Fashion-MNIST~\citep{xiao2017fmnist},
CIFAR-10, CIFAR-100~\citep{krizhevsky2009cifar}, and Food-101~\citep{bossard2014food101}. All target
experiments use the same frozen activation coefficients obtained from the seed-1 source checkpoint:
$a_0=0.2182$, $a_1=0.1193$, $a_2=-0.4237$, $b_1=0.8315$, and $b_2=-0.5918$. Thus, the same
source-learned activation is used across all target architectures and datasets.

\paragraph{Training and evaluation protocol.}
All target experiments use five random seeds, with the same seed configurations across architectures,
datasets, and activation choices. Models are trained for 100 epochs using AdamW, with a learning rate of
$10^{-3}$, weight decay of $0.05$, and a batch size of 256. The learning-rate schedule consists of five
epochs of linear warm-up followed by cosine decay. Label smoothing and dropout are both set to $0.1$.
All activation choices use identical dataset-specific preprocessing and augmentation, without test-time
augmentation.

\ffact{} is compared with ten conventional activation functions: ReLU~\citep{nair2010relu},
LeakyReLU~\citep{maas2013leakyrelu}, GELU~\citep{hendrycks2016gelu}, SiLU~\citep{elfwing2018silu},
Mish~\citep{misra2019mish}, HardSwish~\citep{howard2019mobilenetv3},
Softplus~\citep{dugas2001softplus}, ELU~\citep{clevert2016elu}, SELU~\citep{klambauer2017selu}, and
Tanh~\citep{lecun1998efficient}. For each run, the checkpoint with the highest validation accuracy is
selected and evaluated once on the held-out test set. Results are reported as mean and sample standard
deviation across five seeds. Activation-specific learning-rate and weight-decay tuning results are
provided in Appendix~\ref{app:tuning}.

\subsection{Evolution of the Source-Learned Activation}
\label{sec:evolution}

\begin{figure}[t]
\centering
\includegraphics[width=\linewidth]{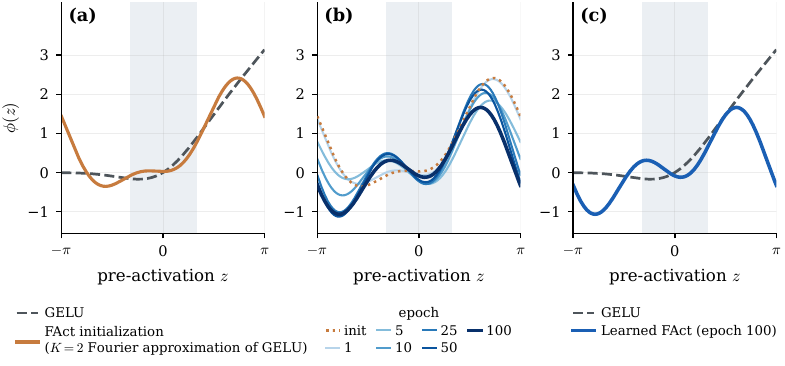}
\caption{\textbf{Evolution of the \fact{} activation during source-network training.} (a) GELU and
its two-harmonic Fourier approximation used to initialize \fact{}. (b) Evolution of \fact{} at selected
training epochs. (c) Final learned activation after 100 epochs compared with GELU. The final curve
corresponds to the seed-1 source checkpoint used in all transfer experiments. The shaded region
indicates the central $98\%$ of source-network pre-activations, pooled across all six MLP blocks on
512 ImageNet-1K validation images.}
\label{fig:evolution}
\end{figure}

\fact{} is initialized using a two-harmonic Fourier approximation of GELU, with all five coefficients
subsequently optimized jointly with the source-network weights. Figure~\ref{fig:evolution} illustrates
the evolution of the shared activation during ImageNet-1K training. The activation progressively departs
from its initial GELU approximation and develops a non-monotonic shape over the course of training. The
final learned function, obtained after 100 epochs, differs substantially from its initialization and is
the activation frozen and transferred to all target networks in the subsequent experiments.

\paragraph{Consistency across source runs.}
Although the transfer experiments reuse the activation from a single source checkpoint, the activation
that \fact{} learns is highly consistent across independent source runs. Training the source network
with five random seeds yields nearly identical coefficients, with mean $\pm$ sample standard deviation
across seeds of $a_0=0.2155\pm0.0030$, $a_1=0.1181\pm0.0038$, $a_2=-0.4205\pm0.0021$,
$b_1=0.8457\pm0.0126$, and $b_2=-0.5990\pm0.0096$.

\subsection{Same-Task Transfer at Source Scale}
\label{sec:same-task}

To examine whether the learned activation remains useful when transferred to a network with the same
architecture and task, we train a freshly initialized \vitsix{} on ImageNet-1K using the frozen \fact{}
activation learned by the source network. The five activation coefficients are obtained from the seed-1
source checkpoint and remain fixed throughout target training. No source-network weights are
transferred. We conduct five independent target runs, all using the same frozen activation, so the
reported variation reflects differences in target initialization and training rather than variability
across independently learned activations.

\begin{table}[ht]
\centering\small
\caption{\textbf{Same-task transfer on ImageNet-1K using a freshly initialized \vitsix{}.} Results
report top-1 accuracy (\%) on the ILSVRC validation set as mean $\pm$ sample standard deviation across
five target runs. \ffact{} uses the activation coefficients learned by the source network;
source-network weights are not transferred, and activation coefficients remain fixed. $\Delta$ denotes
the difference in mean accuracy relative to GELU, in percentage points.}
\label{tab:same-task}
\begin{tabular}{l c c}
\toprule
Activation & Top-1 accuracy (\%) & $\Delta$ vs GELU (\pp) \\
\midrule
GELU & 62.45\,$\pm$\,0.13 & --- \\
\ffact{} & 63.91\,$\pm$\,0.15 & $+1.46$ \\
\bottomrule
\end{tabular}
\end{table}

Table~\ref{tab:same-task} compares \ffact{} with GELU under the same architecture and training
configuration. \ffact{} achieves a mean top-1 accuracy of $63.91\%\pm0.15$, compared with
$62.45\%\pm0.13$ for GELU, an improvement of $1.46$ percentage points. This result provides evidence
that the source-learned activation retains useful properties even when separated from its original
network weights and reused in an independently initialized network performing the same task.

\subsection{Transfer Across Transformer Architectures}
\label{sec:transfer}
\label{sec:swin}
\label{sec:linvit}

\begin{figure}[!t]
\centering
\includegraphics[width=\linewidth]{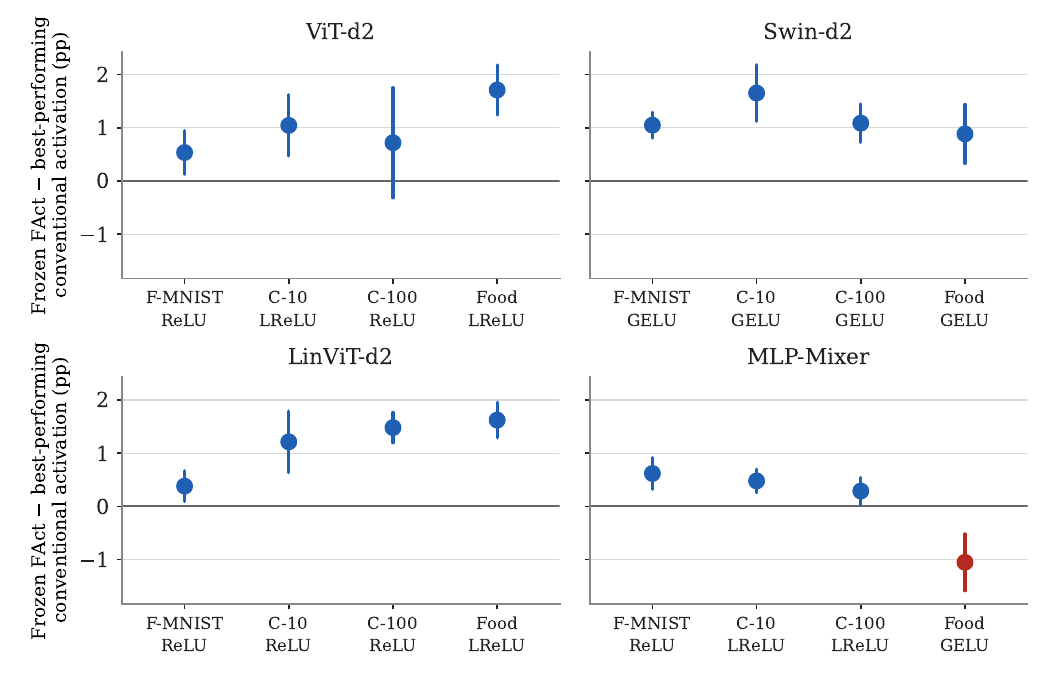}
\caption{\textbf{Accuracy margin of \ffact{} over the best-performing conventional activation across the
sixteen architecture--dataset combinations.} Each panel represents a target architecture. Each point
denotes the mean seed-matched difference in test accuracy (percentage points) across five seeds. Error
bars represent one sample standard deviation of the paired differences. The conventional activation
selected for each comparison is indicated below the corresponding dataset.}
\label{fig:margins}
\end{figure}

Tables~\ref{tab:transfer-shared}--\ref{tab:linvit-main} report the test accuracy of the eleven
activation choices on \vittwo{}, Swin-d2, and LinViT-d2, respectively, under the common training
configuration. Table~\ref{tab:arch} summarizes the performance difference between \ffact{} and the
best-performing conventional activation for each target--dataset combination. Figure~\ref{fig:margins}
illustrates the corresponding accuracy margins in its three transformer panels, alongside those of the
MLP-Mixer target of Section~\ref{sec:mixer}.

\begin{table}[t]
\centering\small
\caption{\textbf{Transfer performance on \vittwo{} with global self-attention under the common training configuration.} Results are mean test accuracy (\%) $\pm$ sample standard deviation across five seeds. Bold and underlining indicate the highest and second-highest mean accuracies, respectively.}
\label{tab:transfer-shared}
\input{tables/table5_transfer_vit}
\end{table}

\begin{table}[t]
\centering\small
\caption{\textbf{Transfer performance on Swin-d2 with shifted-window self-attention under the common training configuration.} Results are mean test accuracy (\%) $\pm$ sample standard deviation across five seeds. Bold and underlining indicate the highest and second-highest mean accuracies, respectively.}
\label{tab:swin-main}
\input{tables/table5_transfer_swin}
\end{table}

\begin{table}[t]
\centering\small
\caption{\textbf{Transfer performance on LinViT-d2 with linear attention under the common training configuration.} Results are mean test accuracy (\%) $\pm$ sample standard deviation across five seeds. Bold and underlining indicate the highest and second-highest mean accuracies, respectively.}
\label{tab:linvit-main}
\input{tables/table5_transfer_linvit}
\end{table}

\begin{table}[t]
\centering\small
\setlength{\tabcolsep}{5pt}
\caption{\textbf{\ffact{} compared with the best-performing conventional activation across the twelve
transformer architecture--dataset combinations under the common training configuration.} The best
conventional activation is selected from the ten baseline functions, excluding \ffact{}. Accuracy values
are mean test accuracy (\%). $\Delta$ denotes the seed-matched mean accuracy difference between \ffact{}
and the selected conventional activation, in percentage points, across five seeds.}
\label{tab:arch}
\input{tables/table5_arch_summary}
\end{table}

Across the three transformer architectures and four datasets, \ffact{} achieves the highest mean test
accuracy in all twelve architecture--dataset combinations. Its improvement over the best-performing
conventional activation ranges from $0.38$ percentage points on LinViT-d2 with Fashion-MNIST to $1.71$
percentage points on \vittwo{} with Food-101. The strongest conventional baseline varies across
architectures: ReLU or LeakyReLU provides the highest mean accuracy among conventional activations for
\vittwo{} and LinViT-d2, whereas GELU performs best across all four Swin-d2 datasets. Importantly, the
same five activation coefficients learned by the source \vitsix{} are used in all target experiments,
without transferring source-network weights or updating the activation during target training. These
results provide empirical evidence that a globally shared nonlinearity learned within one network can
retain its usefulness across independently initialized transformer architectures with different
attention configurations and target tasks.

For reference, we also compare \ffact{} with \fact{} learned directly in each \vittwo{} target under the
same common configuration (Appendix~\ref{app:relearn}). The transferred activation performs comparably to
one learned on the target task: mean differences are within $0.14$ percentage points on Fashion-MNIST and
CIFAR-10, and \ffact{} achieves higher mean accuracy on CIFAR-100 and Food-101.

Additional analyses investigate the effect of the GELU-based initialization and of target-network
depth in Swin targets (Appendix~\ref{app:depth}).

\subsection{Transfer to an MLP-Mixer Architecture}
\label{sec:mixer}

To examine whether the learned activation transfers beyond attention-based architectures, we evaluate
\ffact{} in an independently initialized depth-2 MLP-Mixer~\citep{tolstikhin2021mixer}. Unlike the
transformer architectures, MLP-Mixer uses separate MLPs for token mixing and channel mixing, without
self-attention. We transfer the same five source-learned activation coefficients to both types of MLP
blocks and keep them fixed throughout target training. The comparison includes the same ten
conventional activation functions under the common training configuration.

\begin{table}[!htbp]
\centering\small
\caption{\textbf{Transfer performance on MLP-Mixer} under the common training configuration. Mean test
accuracy (\%) $\pm$ sample s.d.\ across five seeds. Bold indicates the highest mean test accuracy for
each dataset, and underlining indicates the second-highest. The comparison includes \ffact{} and the
same ten conventional activation functions as Tables~\ref{tab:transfer-shared}--\ref{tab:linvit-main}.}
\label{tab:mixer}
\input{tables/table_mixer_common}
\end{table}

\ffact{} achieves the highest mean test accuracy on Fashion-MNIST, CIFAR-10, and CIFAR-100,
outperforming the best-performing conventional activation by $0.62$, $0.48$, and $0.29$ percentage
points, respectively (Table~\ref{tab:mixer} and the MLP-Mixer panel of Figure~\ref{fig:margins}). On
Food-101, GELU performs better by approximately $1.05$ percentage points; this is the only one of the
sixteen target--dataset combinations in which \ffact{} does not achieve the highest mean test accuracy.
These results provide evidence that a source-learned activation can also be reused in an attention-free
architecture without transferring source-network weights or updating its coefficients, although its
performance advantage varies across datasets. After activation-specific hyperparameter tuning, \ffact{}
retains its advantage on Fashion-MNIST and CIFAR-10, whereas HardSwish performs better on CIFAR-100 and
Food-101.

\FloatBarrier
\section{Conclusion}
\label{sec:conclusion}

This work investigated whether a globally shared activation learned jointly with a source network can
be reused independently of the network weights with which it was optimized. We introduced a
learn--freeze--transfer framework and instantiated it using \fact{}, a five-parameter activation
represented by a two-harmonic truncated Fourier series. Initialized using a Fourier approximation of
GELU, its coefficients are optimized jointly with a source Vision Transformer and subsequently frozen
for transfer to independently initialized target networks.

We evaluated the proposed approach through same-task and cross-architecture transfer experiments. In the
same-task setting, \ffact{} improved ImageNet-1K top-1 accuracy by $1.46$ percentage points over GELU
when used in a freshly initialized \vitsix{}. Across four target architectures and four
image-classification datasets, \ffact{} achieved the highest mean test accuracy among eleven activation
choices in 15 of the 16 architecture--dataset combinations under a common training configuration. This
included all twelve combinations involving the three transformer architectures and three of four
MLP-Mixer combinations. Following activation-specific hyperparameter tuning, \ffact{} ranked first in
eleven combinations, second in four, and fourth in the remaining combination. Throughout these
experiments, only the five learned activation coefficients were transferred; source-network weights
were not reused, and the activation remained fixed during target training.

These findings provide empirical evidence that a globally shared nonlinearity learned within one network
can retain its usefulness when transferred to independently initialized networks with different
architectures and target tasks. However, the evaluation uses a single source-learned activation obtained
from one Vision Transformer trained on ImageNet-1K, and the cross-architecture experiments are
restricted to relatively small target networks and image-classification tasks. Further investigation is
needed to establish how activation transferability varies across source architectures, source training
datasets, model scales, and target tasks. Extending this evaluation to convolutional and other network
architectures would help determine the broader applicability of transferable learned nonlinearities.
The mechanism behind the gains also remains open.

{\small
\bibliographystyle{plainnat}
\bibliography{references}
}

\appendix

\section{Experimental Configuration}
\label{app:protocol}

Table~\ref{tab:config} summarizes the architectural configurations and parameter counts of the source
and target networks. The MLP-Mixer consists of two blocks, each comprising a pre-normalized residual
token-mixing MLP (hidden width 38) followed by a pre-normalized residual channel-mixing MLP (hidden
width 304), with an embedding dimension of 76. Classification is performed using globally
average-pooled tokens, without a class token or positional embeddings. All MLPs use a
Linear--activation--dropout--Linear structure, with a single activation module shared across the MLPs
within each network.

The target datasets comprise Fashion-MNIST ($28\times28$), CIFAR-10 and CIFAR-100 ($32\times32$), and
Food-101 (resized to $64\times64$). For each dataset, $5{,}000$ training images are held out using a
fixed random seed to form a validation set shared across experiments. The official test sets are
retained for final evaluation.

\begin{table}[ht]
\centering\footnotesize
\setlength{\tabcolsep}{4pt}
\caption{\textbf{Architectural and dataset configurations of the source and target networks.} Parameter ranges reflect differences across the four target datasets.}
\label{tab:config}
\resizebox{\linewidth}{!}{\input{tables/table_app_config}}
\end{table}

\FloatBarrier
\section{Activation-Specific Hyperparameter Tuning}
\label{app:tuning}
\label{app:tuned}

To examine whether the results under the common training configuration depend on the choice of learning
rate and weight decay, we conduct activation-specific hyperparameter tuning. For each
activation--architecture--dataset combination, we evaluate 15 configurations obtained from five learning
rates ($3\times10^{-4}$, $10^{-3}$, $3\times10^{-3}$, $10^{-2}$, and $3\times10^{-2}$) and three
weight-decay values ($0$, $0.05$, and $0.2$). Each configuration is evaluated using a 30-epoch,
single-seed proxy run. The selected configuration is then trained for 100 epochs using five random
seeds. The same search budget is applied to all eleven activation functions across the four target
architectures and four datasets.

\begin{table}[ht]
\centering\small
\caption{\textbf{\ffact{} compared with the best-performing conventional activation under
activation-specific hyperparameter tuning.} Results are mean test accuracy (\%) $\pm$ sample standard
deviation across five seeds for both \ffact{} and the best-performing conventional activation.
$\Delta$ denotes the difference in mean accuracy between \ffact{} and the selected conventional
activation, in percentage points.}
\label{tab:tuned-summary}
\input{tables/table_app_tuned}
\end{table}

Table~\ref{tab:tuned-summary} compares \ffact{} with the best-performing conventional activation under
the tuned configurations. \ffact{} achieves the highest mean accuracy in 11 of the 16
architecture--dataset combinations, ranks second in four, and fourth in the remaining combination.
Across the three transformer architectures, it ranks first in nine of the twelve combinations, with GELU
achieving higher accuracy in the other three. For the MLP-Mixer, \ffact{} retains the highest accuracy on
Fashion-MNIST and CIFAR-10, whereas HardSwish performs better on CIFAR-100 and Food-101.

\begin{figure}[!tp]
\centering
\includegraphics[width=0.92\linewidth]{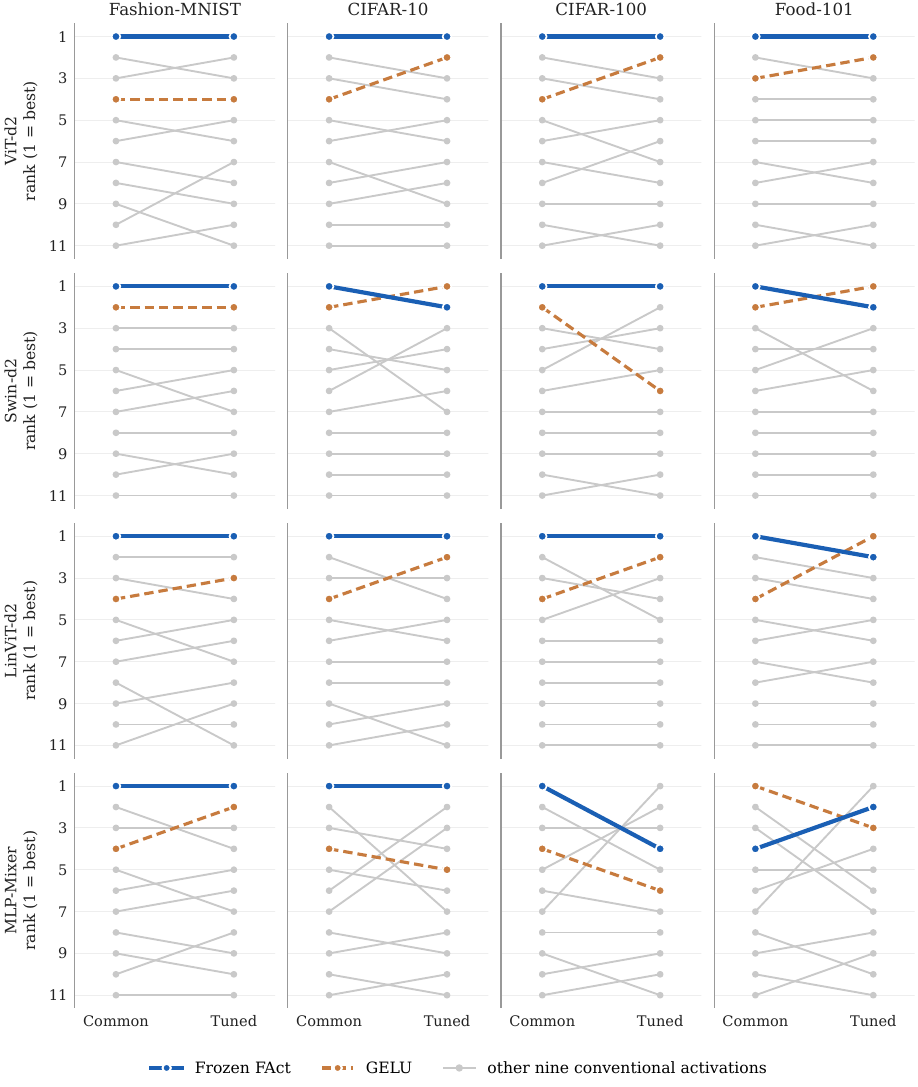}
\caption{\textbf{Activation rankings before and after activation-specific hyperparameter tuning.} Each
panel shows the ranking of eleven activation choices based on mean test accuracy across five seeds. The
left and right positions represent rankings under the common and tuned configurations, respectively
(rank 1 = highest accuracy). Lines connect the rankings of the same activation across the two
configurations. \ffact{} and GELU are highlighted; the remaining nine conventional activations are
shown in grey.}
\label{fig:tuning-ranks}
\end{figure}

Figure~\ref{fig:tuning-ranks} illustrates the changes in activation rankings between the common and tuned
configurations. Although activation-specific tuning changes the relative rankings, \ffact{} retains the
highest mean accuracy in most of the evaluated combinations. These results indicate that its observed
advantage is not eliminated by activation-specific hyperparameter optimization, although its relative
performance remains dependent on the target architecture and dataset.

\FloatBarrier
\section{Depth Sensitivity and the Effect of GELU-Based Initialization}
\label{app:depth}

This experiment examines whether the performance of \ffact{} can be attributed to the learned activation
coefficients rather than merely to the Fourier approximation of GELU, and how its transfer performance
varies with target-network depth. We use Swin Transformer targets because GELU is the strongest
conventional activation across all four datasets in the main experiments (Table~\ref{tab:swin-main}),
providing a consistent reference for both comparisons. The findings are therefore specific to the
evaluated Swin configurations.

\subsection{Experimental Design}

We vary the depth of the target Swin Transformer from one to six blocks while keeping the other
architectural settings fixed. All models use a single stage with an embedding dimension of 64, four
attention heads, an MLP ratio of 4, a patch size of 4, and a window size of 4. Blocks alternate between
regular-window and shifted-window attention, beginning with a regular-window block followed by a
shifted-window block with a shift of two tokens. No patch merging is applied, keeping the token grid
unchanged across depths. Thus, the primary architectural variable is the number of blocks and the
corresponding number of activation application sites.

We compare three activation configurations: (1) GELU; (2) Frozen GELU-init, the two-harmonic Fourier
approximation of GELU obtained using the initialization in Eq.~\ref{eq:init}, with its coefficients fixed
throughout training; and (3) \ffact{}, using the coefficients learned by the ImageNet-1K source ViT and
transferred without further updates. All target networks are independently initialized and trained using
five random seeds under the common training configuration described in Section~\ref{sec:setup}. Neither
frozen configuration uses source-network weights.

\begin{table}[ht]
\centering\small
\caption{\textbf{Test accuracy across Swin Transformer target depths.} Results are mean accuracy (\%)
$\pm$ sample standard deviation across five seeds. The three configurations are GELU, Frozen
GELU-init and \ffact{}. Bold indicates the highest mean accuracy at each dataset--depth combination.}
\label{tab:depth}
\input{tables/table_app_depth_swin}
\end{table}

\begin{figure}[ht]
\centering
\includegraphics[width=\linewidth]{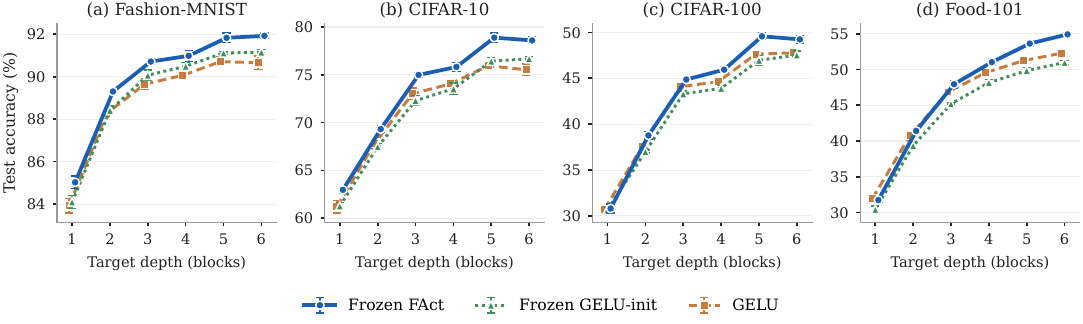}
\caption{\textbf{Effect of target-network depth on activation performance.} Test accuracy of GELU,
Frozen GELU-init and \ffact{} across Swin Transformer target depths from one to six blocks on
Fashion-MNIST, CIFAR-10, CIFAR-100 and Food-101. Points represent mean test accuracy across five
seeds, and error bars indicate one sample standard deviation. All target networks are independently
initialized and trained with their respective activation configurations.}
\label{fig:depth}
\end{figure}

\subsection{Results}

\subsubsection{Effect of the GELU-Based Fourier Initialization}

We first compare Frozen GELU-init with GELU to examine whether the fixed Fourier approximation alone
provides an advantage (Table~\ref{tab:depth}). The results show that its performance is dataset-dependent
rather than consistently better than GELU. Although Frozen GELU-init achieves higher mean accuracy at all
six depths on Fashion-MNIST, its advantage is inconsistent on CIFAR-10 and absent on CIFAR-100 and
Food-101. Thus, replacing GELU with its fixed Fourier approximation does not, by itself, explain the
improvements observed with \ffact{}.

\subsubsection{Sensitivity to Target-Network Depth}

We next examine how the performance of the transferred activation varies with target-network depth
(Table~\ref{tab:depth} and Figure~\ref{fig:depth}). \ffact{} achieves the highest mean accuracy among the
three configurations at every depth on Fashion-MNIST, CIFAR-10, and CIFAR-100, and at five of the six
depths on Food-101.

Notably, \ffact{} outperforms Frozen GELU-init across all 24 dataset--depth combinations, with
improvements ranging from $0.09$ to $3.94$ percentage points. This indicates that the gains are not
explained solely by the Fourier representation of GELU; the source-learned coefficients contribute
additional benefits under the evaluated settings. The advantage over GELU also varies with dataset and
depth, becoming more pronounced at greater depths on CIFAR-10, CIFAR-100, and Food-101.

Overall, these results show that the transferred activation remains effective across the evaluated Swin
target depths, while its performance advantage depends on the target dataset and network depth.

\section{Comparison with Activation Learning in the Target Network}
\label{app:relearn}

To compare activation transfer with learning an activation directly for the target task, we train
\fact{} jointly with each \vittwo{} target network (Learnable \fact{}), using the same Fourier
parameterization and GELU-based initialization as the source network. We compare this configuration with
\ffact{} transferred from the ImageNet-1K source ViT. Both configurations use the common training
protocol and five identical random seeds. The \ffact{} results correspond to the same runs reported in
Table~\ref{tab:transfer-shared}.

\begin{table}[h]
\centering
\caption{Comparison of \ffact{} transferred from the source \vitsix{} and Learnable \fact{} optimized
within the target \vittwo{}. Results report mean test accuracy (\%) $\pm$ sample standard deviation
across five seeds. $\Delta$ denotes the difference in mean accuracy (Frozen $-$ Learnable), in
percentage points.}
\label{tab:relearn}
\begin{tabular}{lccc}
\toprule
Dataset & Learnable \fact{} & \ffact{} & $\Delta$ (pp) \\
\midrule
Fashion-MNIST & $91.08 \pm 0.20$ & $90.94 \pm 0.10$ & $-0.14$ \\
CIFAR-10      & $75.56 \pm 0.21$ & $75.56 \pm 0.35$ & $0.00$ \\
CIFAR-100     & $46.73 \pm 0.32$ & $47.67 \pm 0.48$ & $+0.94$ \\
Food-101      & $35.20 \pm 0.40$ & $35.83 \pm 0.36$ & $+0.63$ \\
\bottomrule
\end{tabular}
\end{table}

Table~\ref{tab:relearn} summarizes the comparison. On Fashion-MNIST and CIFAR-10, the mean accuracy
differences between the two configurations are $-0.14$ and $0.00$ percentage points, respectively. On
CIFAR-100 and Food-101, \ffact{} achieves higher mean accuracies by $0.94$ and $0.63$ percentage points,
respectively.

These results indicate that, in the evaluated \vittwo{} settings, transferring the source-learned
activation can achieve performance comparable to or higher than learning the activation directly within
the target network. This comparison is limited to \vittwo{} and does not establish that the two
approaches are equivalent across architectures or datasets.

\end{document}

%% file: tables/table5_transfer_vit.tex
\begin{tabular}{l cccc}
\toprule
Activation & Fashion-MNIST & CIFAR-10 & CIFAR-100 & Food-101 \\
\midrule
\ffact{} & \textbf{90.94\,\boldmath$\pm$\unboldmath{}\,0.10} & \textbf{75.56\,\boldmath$\pm$\unboldmath{}\,0.35} & \textbf{47.67\,\boldmath$\pm$\unboldmath{}\,0.48} & \textbf{35.83\,\boldmath$\pm$\unboldmath{}\,0.36} \\
\midrule
ReLU & \underline{90.40\,$\pm$\,0.46} & 74.50\,$\pm$\,0.46 & \underline{46.95\,$\pm$\,0.63} & 34.07\,$\pm$\,0.22 \\
LeakyReLU & 90.27\,$\pm$\,0.27 & \underline{74.51\,$\pm$\,0.26} & 46.80\,$\pm$\,0.21 & \underline{34.12\,$\pm$\,0.21} \\
\addlinespace[2pt]
GELU & 89.09\,$\pm$\,0.24 & 72.46\,$\pm$\,0.29 & 46.30\,$\pm$\,0.24 & 34.10\,$\pm$\,0.39 \\
SiLU & 87.96\,$\pm$\,0.57 & 70.47\,$\pm$\,0.39 & 44.60\,$\pm$\,0.34 & 32.51\,$\pm$\,0.54 \\
Mish & 87.91\,$\pm$\,0.35 & 70.62\,$\pm$\,0.49 & 44.85\,$\pm$\,0.31 & 32.69\,$\pm$\,0.32 \\
HardSwish & 87.53\,$\pm$\,0.58 & 68.01\,$\pm$\,0.15 & 42.13\,$\pm$\,0.33 & 31.10\,$\pm$\,0.72 \\
\addlinespace[2pt]
Softplus & 86.27\,$\pm$\,0.37 & 68.51\,$\pm$\,0.41 & 40.44\,$\pm$\,0.80 & 31.84\,$\pm$\,0.22 \\
ELU & 86.32\,$\pm$\,0.09 & 66.25\,$\pm$\,0.19 & 38.44\,$\pm$\,0.68 & 29.36\,$\pm$\,0.57 \\
SELU & 86.24\,$\pm$\,0.51 & 65.29\,$\pm$\,0.50 & 37.12\,$\pm$\,0.26 & 28.71\,$\pm$\,0.24 \\
Tanh & 86.61\,$\pm$\,0.20 & 65.14\,$\pm$\,0.32 & 37.24\,$\pm$\,0.20 & 28.67\,$\pm$\,0.40 \\
\bottomrule
\end{tabular}

%% file: tables/table5_transfer_swin.tex
\begin{tabular}{l cccc}
\toprule
Activation & Fashion-MNIST & CIFAR-10 & CIFAR-100 & Food-101 \\
\midrule
\ffact{} & \textbf{89.33\,\boldmath$\pm$\unboldmath{}\,0.11} & \textbf{69.47\,\boldmath$\pm$\unboldmath{}\,0.31} & \textbf{38.34\,\boldmath$\pm$\unboldmath{}\,0.34} & \textbf{41.32\,\boldmath$\pm$\unboldmath{}\,0.26} \\
\midrule
ReLU & 88.28\,$\pm$\,0.28 & 67.31\,$\pm$\,0.33 & 36.87\,$\pm$\,0.44 & 39.20\,$\pm$\,0.55 \\
LeakyReLU & 88.27\,$\pm$\,0.15 & 67.56\,$\pm$\,0.41 & 36.85\,$\pm$\,0.31 & 39.20\,$\pm$\,0.37 \\
\addlinespace[2pt]
GELU & \underline{88.29\,$\pm$\,0.22} & \underline{67.82\,$\pm$\,0.31} & \underline{37.25\,$\pm$\,0.27} & \underline{40.43\,$\pm$\,0.38} \\
SiLU & 87.69\,$\pm$\,0.18 & 66.21\,$\pm$\,0.35 & 36.48\,$\pm$\,0.44 & 39.01\,$\pm$\,0.54 \\
Mish & 87.60\,$\pm$\,0.09 & 66.44\,$\pm$\,0.32 & 36.97\,$\pm$\,0.68 & 39.34\,$\pm$\,0.32 \\
HardSwish & 87.33\,$\pm$\,0.24 & 65.60\,$\pm$\,0.27 & 36.14\,$\pm$\,0.32 & 38.17\,$\pm$\,0.75 \\
\addlinespace[2pt]
Softplus & 85.73\,$\pm$\,0.27 & 62.61\,$\pm$\,0.26 & 32.39\,$\pm$\,0.40 & 35.67\,$\pm$\,0.82 \\
ELU & 85.55\,$\pm$\,0.39 & 61.03\,$\pm$\,0.65 & 31.55\,$\pm$\,0.35 & 32.66\,$\pm$\,0.58 \\
SELU & 83.27\,$\pm$\,0.40 & 54.81\,$\pm$\,0.68 & 25.71\,$\pm$\,0.29 & 28.75\,$\pm$\,0.43 \\
Tanh & 85.69\,$\pm$\,0.37 & 57.63\,$\pm$\,0.90 & 28.06\,$\pm$\,0.41 & 30.35\,$\pm$\,0.43 \\
\bottomrule
\end{tabular}

%% file: tables/table5_transfer_linvit.tex
\begin{tabular}{l cccc}
\toprule
Activation & Fashion-MNIST & CIFAR-10 & CIFAR-100 & Food-101 \\
\midrule
\ffact{} & \textbf{90.67\,\boldmath$\pm$\unboldmath{}\,0.24} & \textbf{75.15\,\boldmath$\pm$\unboldmath{}\,0.38} & \textbf{48.05\,\boldmath$\pm$\unboldmath{}\,0.19} & \textbf{36.12\,\boldmath$\pm$\unboldmath{}\,0.34} \\
\midrule
ReLU & \underline{90.29\,$\pm$\,0.16} & \underline{73.94\,$\pm$\,0.48} & \underline{46.57\,$\pm$\,0.36} & 34.40\,$\pm$\,0.27 \\
LeakyReLU & 90.18\,$\pm$\,0.23 & 73.89\,$\pm$\,0.30 & 46.35\,$\pm$\,0.29 & \underline{34.50\,$\pm$\,0.29} \\
\addlinespace[2pt]
GELU & 89.07\,$\pm$\,0.22 & 72.13\,$\pm$\,0.31 & 45.47\,$\pm$\,0.32 & 33.88\,$\pm$\,0.63 \\
SiLU & 87.89\,$\pm$\,0.52 & 70.05\,$\pm$\,0.31 & 43.29\,$\pm$\,0.46 & 32.49\,$\pm$\,0.13 \\
Mish & 88.06\,$\pm$\,0.25 & 70.23\,$\pm$\,0.28 & 43.66\,$\pm$\,0.63 & 32.70\,$\pm$\,0.17 \\
HardSwish & 87.15\,$\pm$\,0.40 & 68.35\,$\pm$\,0.47 & 41.16\,$\pm$\,0.62 & 31.35\,$\pm$\,0.31 \\
\addlinespace[2pt]
Softplus & 85.48\,$\pm$\,0.26 & 64.30\,$\pm$\,0.98 & 38.58\,$\pm$\,0.48 & 31.47\,$\pm$\,0.28 \\
ELU & 86.02\,$\pm$\,0.24 & 65.63\,$\pm$\,0.87 & 37.09\,$\pm$\,0.24 & 29.24\,$\pm$\,0.34 \\
SELU & 85.95\,$\pm$\,0.43 & 65.07\,$\pm$\,0.62 & 35.56\,$\pm$\,0.55 & 27.91\,$\pm$\,0.38 \\
Tanh & 86.28\,$\pm$\,0.50 & 64.85\,$\pm$\,0.88 & 35.57\,$\pm$\,0.51 & 28.04\,$\pm$\,0.11 \\
\bottomrule
\end{tabular}

%% file: tables/table5_arch_summary.tex
\begin{tabular}{l l c l c}
\toprule
Target & Dataset & \ffact{} & \begin{tabular}[b]{@{}l@{}}Best-performing\\conventional activation\end{tabular} & $\Delta$ (\pp) \\
\midrule
\vittwo{} & Fashion-MNIST & 90.94 & ReLU (90.40) & $+0.53$ \\
 & CIFAR-10 & 75.56 & LeakyReLU (74.51) & $+1.04$ \\
 & CIFAR-100 & 47.67 & ReLU (46.95) & $+0.72$ \\
 & Food-101 & 35.83 & LeakyReLU (34.12) & $+1.71$ \\
\addlinespace[3pt]
Swin-d2 & Fashion-MNIST & 89.33 & GELU (88.29) & $+1.05$ \\
 & CIFAR-10 & 69.47 & GELU (67.82) & $+1.65$ \\
 & CIFAR-100 & 38.34 & GELU (37.25) & $+1.09$ \\
 & Food-101 & 41.32 & GELU (40.43) & $+0.88$ \\
\addlinespace[3pt]
LinViT-d2 & Fashion-MNIST & 90.67 & ReLU (90.29) & $+0.38$ \\
 & CIFAR-10 & 75.15 & ReLU (73.94) & $+1.21$ \\
 & CIFAR-100 & 48.05 & ReLU (46.57) & $+1.48$ \\
 & Food-101 & 36.12 & LeakyReLU (34.50) & $+1.62$ \\
\bottomrule
\end{tabular}

%% file: tables/table_mixer_common.tex
\begin{tabular}{l cccc}
\toprule
Activation & Fashion-MNIST & CIFAR-10 & CIFAR-100 & Food-101 \\
\midrule
\ffact{} & \textbf{91.42\,\boldmath$\pm$\unboldmath{}\,0.18} & \textbf{75.99\,\boldmath$\pm$\unboldmath{}\,0.16} & \textbf{47.49\,\boldmath$\pm$\unboldmath{}\,0.38} & 35.09\,$\pm$\,0.22 \\
\midrule
ReLU & \underline{90.80\,$\pm$\,0.22} & 75.47\,$\pm$\,0.35 & 47.08\,$\pm$\,0.46 & 34.92\,$\pm$\,0.35 \\
LeakyReLU & 90.64\,$\pm$\,0.39 & \underline{75.51\,$\pm$\,0.13} & \underline{47.20\,$\pm$\,0.48} & 34.94\,$\pm$\,0.40 \\
\addlinespace[2pt]
GELU & 89.84\,$\pm$\,0.13 & 75.32\,$\pm$\,0.48 & 45.99\,$\pm$\,0.60 & \textbf{36.15\,\boldmath$\pm$\unboldmath{}\,0.47} \\
SiLU & 88.90\,$\pm$\,0.29 & 73.75\,$\pm$\,0.30 & 45.07\,$\pm$\,0.83 & 35.15\,$\pm$\,0.30 \\
Mish & 88.94\,$\pm$\,0.25 & 74.03\,$\pm$\,0.30 & 45.17\,$\pm$\,0.58 & \underline{35.25\,$\pm$\,0.25} \\
HardSwish & 87.81\,$\pm$\,0.41 & 72.78\,$\pm$\,0.42 & 44.05\,$\pm$\,0.78 & 34.39\,$\pm$\,0.26 \\
\addlinespace[2pt]
Softplus & 86.98\,$\pm$\,0.42 & 68.26\,$\pm$\,1.06 & 35.33\,$\pm$\,1.84 & 32.70\,$\pm$\,0.12 \\
ELU & 86.52\,$\pm$\,0.22 & 70.19\,$\pm$\,0.56 & 41.27\,$\pm$\,0.78 & 30.82\,$\pm$\,0.38 \\
SELU & 86.02\,$\pm$\,0.22 & 64.97\,$\pm$\,0.22 & 36.07\,$\pm$\,0.46 & 24.90\,$\pm$\,0.25 \\
Tanh & 86.53\,$\pm$\,0.18 & 65.42\,$\pm$\,0.56 & 36.19\,$\pm$\,0.43 & 29.97\,$\pm$\,0.29 \\
\bottomrule
\end{tabular}

%% file: tables/table_app_config.tex
\begin{tabular}{l l l l l l}
\toprule
& Source & \multicolumn{4}{c}{Targets} \\
\cmidrule(lr){2-2}\cmidrule(lr){3-6}
& \vitsix{} & \vittwo{} & Swin-d2 & LinViT-d2 & MLP-Mixer \\
\midrule
Token mixing & global softmax & global softmax & shifted window (4) & linear, $\mathrm{elu}{+}1$ & MLP \\
Depth / width / heads & 6 / 192 / 6 & 2 / 64 / 4 & 2 / 64 / 4 & 2 / 64 / 4 & 2 / 76 / -- \\
MLP hidden width & 768 & 256 & 256 & 256 & 38 token / 304 channel \\
Patch size & 16 & 4 (Food-101: 8) & 4 & 4 (Food-101: 8) & 4 (Food-101: 8) \\
Parameters & 3{,}048{,}232 & 105{,}098--123{,}237 & 102{,}354--110{,}317 & 105{,}098--123{,}237 & 103{,}620--126{,}313 \\
\addlinespace[2pt]
Data & ImageNet-1K, $224^2$ & \multicolumn{4}{l}{Fashion-MNIST $28^2$, CIFAR-10/100 $32^2$, Food-101 $64^2$} \\
Validation split & 10{,}000 train images & \multicolumn{4}{l}{5{,}000 train images per dataset, fixed} \\
Test split & ILSVRC val (50{,}000) & \multicolumn{4}{l}{official test set} \\
\bottomrule
\end{tabular}

%% file: tables/table_app_tuned.tex
\begin{tabular}{l l c l c c}
\toprule
& & & \multicolumn{2}{c}{Best-performing conventional} & \\
\cmidrule(lr){4-5}
Target & Dataset & \ffact{} & Activation & Accuracy & $\Delta$ (\pp) \\
\midrule
\vittwo{} & Fashion-MNIST & 91.17\,$\pm$\,0.17 & LeakyReLU & 90.69\,$\pm$\,0.31 & $+0.48$ \\
 & CIFAR-10 & 76.34\,$\pm$\,0.69 & GELU & 74.91\,$\pm$\,0.13 & $+1.43$ \\
 & CIFAR-100 & 48.33\,$\pm$\,0.48 & GELU & 47.06\,$\pm$\,0.30 & $+1.27$ \\
 & Food-101 & 39.15\,$\pm$\,0.49 & GELU & 37.13\,$\pm$\,0.41 & $+2.01$ \\
\addlinespace[3pt]
Swin-d2 & Fashion-MNIST & 90.51\,$\pm$\,0.23 & GELU & 90.13\,$\pm$\,0.17 & $+0.37$ \\
 & CIFAR-10 & 72.57\,$\pm$\,0.57 & GELU & 73.07\,$\pm$\,0.22 & ${-}0.50$ \\
 & CIFAR-100 & 44.13\,$\pm$\,0.54 & LeakyReLU & 42.18\,$\pm$\,0.59 & $+1.95$ \\
 & Food-101 & 46.51\,$\pm$\,0.40 & GELU & 46.60\,$\pm$\,0.86 & ${-}0.09$ \\
\addlinespace[3pt]
LinViT-d2 & Fashion-MNIST & 91.39\,$\pm$\,0.15 & ReLU & 90.76\,$\pm$\,0.27 & $+0.63$ \\
 & CIFAR-10 & 76.38\,$\pm$\,0.45 & GELU & 74.44\,$\pm$\,0.44 & $+1.95$ \\
 & CIFAR-100 & 48.51\,$\pm$\,0.41 & GELU & 48.18\,$\pm$\,0.36 & $+0.33$ \\
 & Food-101 & 38.34\,$\pm$\,0.25 & GELU & 38.70\,$\pm$\,0.64 & ${-}0.36$ \\
\addlinespace[3pt]
MLP-Mixer & Fashion-MNIST & 91.69\,$\pm$\,0.15 & GELU & 90.93\,$\pm$\,0.29 & $+0.76$ \\
 & CIFAR-10 & 77.89\,$\pm$\,0.66 & SiLU & 77.70\,$\pm$\,0.44 & $+0.19$ \\
 & CIFAR-100 & 49.01\,$\pm$\,0.56 & HardSwish & 51.48\,$\pm$\,0.91 & ${-}2.47$ \\
 & Food-101 & 42.48\,$\pm$\,0.10 & HardSwish & 43.27\,$\pm$\,1.21 & ${-}0.79$ \\
\bottomrule
\end{tabular}

%% file: tables/table_app_depth_swin.tex
\begin{tabular}{l c c c c}
\toprule
Dataset & Depth & GELU & Frozen GELU-init & \ffact{} \\
\midrule
Fashion-MNIST & 1 & 83.93\,$\pm$\,0.35 & 84.10\,$\pm$\,0.32 & \textbf{85.03}\,$\pm$\,0.30 \\
 & 2 & 88.28\,$\pm$\,0.18 & 88.41\,$\pm$\,0.20 & \textbf{89.30}\,$\pm$\,0.06 \\
 & 3 & 89.65\,$\pm$\,0.26 & 90.07\,$\pm$\,0.24 & \textbf{90.71}\,$\pm$\,0.13 \\
 & 4 & 90.06\,$\pm$\,0.06 & 90.49\,$\pm$\,0.18 & \textbf{90.98}\,$\pm$\,0.26 \\
 & 5 & 90.71\,$\pm$\,0.24 & 91.11\,$\pm$\,0.07 & \textbf{91.83}\,$\pm$\,0.23 \\
 & 6 & 90.66\,$\pm$\,0.33 & 91.15\,$\pm$\,0.13 & \textbf{91.92}\,$\pm$\,0.13 \\
\addlinespace[3pt]
CIFAR-10 & 1 & 61.19\,$\pm$\,0.67 & 61.27\,$\pm$\,0.39 & \textbf{62.96}\,$\pm$\,0.35 \\
 & 2 & 67.83\,$\pm$\,0.19 & 67.47\,$\pm$\,0.35 & \textbf{69.31}\,$\pm$\,0.34 \\
 & 3 & 73.01\,$\pm$\,0.62 & 72.30\,$\pm$\,0.52 & \textbf{74.98}\,$\pm$\,0.28 \\
 & 4 & 74.05\,$\pm$\,0.39 & 73.53\,$\pm$\,0.62 & \textbf{75.82}\,$\pm$\,0.47 \\
 & 5 & 75.94\,$\pm$\,0.28 & 76.43\,$\pm$\,0.38 & \textbf{78.89}\,$\pm$\,0.48 \\
 & 6 & 75.53\,$\pm$\,0.62 & 76.68\,$\pm$\,0.25 & \textbf{78.60}\,$\pm$\,0.34 \\
\addlinespace[3pt]
CIFAR-100 & 1 & 30.72\,$\pm$\,0.42 & 30.71\,$\pm$\,0.18 & \textbf{30.80}\,$\pm$\,0.48 \\
 & 2 & 37.61\,$\pm$\,0.35 & 37.00\,$\pm$\,0.17 & \textbf{38.78}\,$\pm$\,0.40 \\
 & 3 & 44.09\,$\pm$\,0.37 & 43.31\,$\pm$\,0.31 & \textbf{44.87}\,$\pm$\,0.19 \\
 & 4 & 44.63\,$\pm$\,0.21 & 43.91\,$\pm$\,0.29 & \textbf{45.94}\,$\pm$\,0.31 \\
 & 5 & 47.64\,$\pm$\,0.27 & 46.96\,$\pm$\,0.51 & \textbf{49.58}\,$\pm$\,0.36 \\
 & 6 & 47.78\,$\pm$\,0.47 & 47.54\,$\pm$\,0.42 & \textbf{49.25}\,$\pm$\,0.40 \\
\addlinespace[3pt]
Food-101 & 1 & \textbf{31.93}\,$\pm$\,0.24 & 30.43\,$\pm$\,0.51 & 31.74\,$\pm$\,0.33 \\
 & 2 & 40.79\,$\pm$\,0.34 & 39.35\,$\pm$\,0.35 & \textbf{41.41}\,$\pm$\,0.34 \\
 & 3 & 46.85\,$\pm$\,0.16 & 45.20\,$\pm$\,0.14 & \textbf{47.95}\,$\pm$\,0.55 \\
 & 4 & 49.57\,$\pm$\,0.39 & 48.16\,$\pm$\,0.44 & \textbf{51.03}\,$\pm$\,0.43 \\
 & 5 & 51.26\,$\pm$\,0.30 & 49.85\,$\pm$\,0.44 & \textbf{53.63}\,$\pm$\,0.39 \\
 & 6 & 52.27\,$\pm$\,0.41 & 50.98\,$\pm$\,0.26 & \textbf{54.92}\,$\pm$\,0.19 \\
\bottomrule
\end{tabular}

%% file: LeFT-nips.bbl
\begin{thebibliography}{37}
\providecommand{\natexlab}[1]{#1}
\providecommand{\url}[1]{\texttt{#1}}
\expandafter\ifx\csname urlstyle\endcsname\relax
  \providecommand{\doi}[1]{doi: #1}\else
  \providecommand{\doi}{doi: \begingroup \urlstyle{rm}\Url}\fi

\bibitem[Agostinelli et~al.(2015)Agostinelli, Hoffman, Sadowski, and
  Baldi]{agostinelli2015apl}
Forest Agostinelli, Matthew Hoffman, Peter Sadowski, and Pierre Baldi.
\newblock Learning activation functions to improve deep neural networks.
\newblock In \emph{ICLR Workshop}, 2015.

\bibitem[Bingham and Miikkulainen(2022)]{bingham2022pangaea}
Garrett Bingham and Risto Miikkulainen.
\newblock Discovering parametric activation functions.
\newblock \emph{Neural Networks}, 148:\penalty0 48--65, 2022.

\bibitem[Bossard et~al.(2014)Bossard, Guillaumin, and
  Van~Gool]{bossard2014food101}
Lukas Bossard, Matthieu Guillaumin, and Luc Van~Gool.
\newblock Food-101 -- mining discriminative components with random forests.
\newblock In \emph{European Conference on Computer Vision (ECCV)}, pages
  446--461, 2014.

\bibitem[Clevert et~al.(2016)Clevert, Unterthiner, and
  Hochreiter]{clevert2016elu}
Djork-Arn{\'e} Clevert, Thomas Unterthiner, and Sepp Hochreiter.
\newblock Fast and accurate deep network learning by exponential linear units
  ({ELU}s).
\newblock In \emph{International Conference on Learning Representations
  (ICLR)}, 2016.

\bibitem[Delfosse et~al.(2024)Delfosse, Schramowski, Mundt, Molina, and
  Kersting]{delfosse2024rational}
Quentin Delfosse, Patrick Schramowski, Martin Mundt, Alejandro Molina, and
  Kristian Kersting.
\newblock Adaptive rational activations to boost deep reinforcement learning.
\newblock In \emph{International Conference on Learning Representations
  (ICLR)}, 2024.
\newblock Earlier version: Recurrent Rational Networks, arXiv:2102.09407.

\bibitem[Deng et~al.(2009)Deng, Dong, Socher, Li, Li, and
  Fei-Fei]{deng2009imagenet}
Jia Deng, Wei Dong, Richard Socher, Li-Jia Li, Kai Li, and Li~Fei-Fei.
\newblock {I}mage{N}et: A large-scale hierarchical image database.
\newblock In \emph{IEEE Conference on Computer Vision and Pattern Recognition
  (CVPR)}, pages 248--255, 2009.

\bibitem[Dosovitskiy et~al.(2021)Dosovitskiy, Beyer, Kolesnikov, Weissenborn,
  Zhai, Unterthiner, Dehghani, Minderer, Heigold, Gelly, Uszkoreit, and
  Houlsby]{dosovitskiy2021vit}
Alexey Dosovitskiy, Lucas Beyer, Alexander Kolesnikov, Dirk Weissenborn,
  Xiaohua Zhai, Thomas Unterthiner, Mostafa Dehghani, Matthias Minderer, Georg
  Heigold, Sylvain Gelly, Jakob Uszkoreit, and Neil Houlsby.
\newblock An image is worth 16x16 words: Transformers for image recognition at
  scale.
\newblock In \emph{International Conference on Learning Representations
  (ICLR)}, 2021.

\bibitem[Dugas et~al.(2001)Dugas, Bengio, B{\'e}lisle, Nadeau, and
  Garcia]{dugas2001softplus}
Charles Dugas, Yoshua Bengio, Fran{\c c}ois B{\'e}lisle, Claude Nadeau, and
  Ren{\'e} Garcia.
\newblock Incorporating second-order functional knowledge for better option
  pricing.
\newblock In \emph{Advances in Neural Information Processing Systems (NIPS)},
  pages 472--478, 2001.

\bibitem[Elfwing et~al.(2018)Elfwing, Uchibe, and Doya]{elfwing2018silu}
Stefan Elfwing, Eiji Uchibe, and Kenji Doya.
\newblock Sigmoid-weighted linear units for neural network function
  approximation in reinforcement learning.
\newblock \emph{Neural Networks}, 107:\penalty0 3--11, 2018.

\bibitem[He et~al.(2015)He, Zhang, Ren, and Sun]{he2015prelu}
Kaiming He, Xiangyu Zhang, Shaoqing Ren, and Jian Sun.
\newblock Delving deep into rectifiers: Surpassing human-level performance on
  {I}mage{N}et classification.
\newblock In \emph{IEEE International Conference on Computer Vision (ICCV)},
  pages 1026--1034, 2015.

\bibitem[Hendrycks and Gimpel(2016)]{hendrycks2016gelu}
Dan Hendrycks and Kevin Gimpel.
\newblock {G}aussian error linear units ({GELU}s).
\newblock \emph{arXiv:1606.08415}, 2016.

\bibitem[Hinton et~al.(2015)Hinton, Vinyals, and Dean]{hinton2015distilling}
Geoffrey Hinton, Oriol Vinyals, and Jeff Dean.
\newblock Distilling the knowledge in a neural network.
\newblock \emph{arXiv preprint arXiv:1503.02531}, 2015.

\bibitem[Howard et~al.(2019)Howard, Sandler, Chu, Chen, Chen, Tan, Wang, Zhu,
  Pang, Vasudevan, Le, and Adam]{howard2019mobilenetv3}
Andrew Howard, Mark Sandler, Grace Chu, Liang-Chieh Chen, Bo~Chen, Mingxing
  Tan, Weijun Wang, Yukun Zhu, Ruoming Pang, Vijay Vasudevan, Quoc~V. Le, and
  Hartwig Adam.
\newblock Searching for {M}obile{N}et{V}3.
\newblock In \emph{IEEE/CVF International Conference on Computer Vision
  (ICCV)}, pages 1314--1324, 2019.

\bibitem[Jagtap et~al.(2020{\natexlab{a}})Jagtap, Kawaguchi, and
  Karniadakis]{jagtap2020gaaf}
Ameya~D. Jagtap, Kenji Kawaguchi, and George~Em Karniadakis.
\newblock Adaptive activation functions accelerate convergence in deep and
  physics-informed neural networks.
\newblock \emph{Journal of Computational Physics}, 404:\penalty0 109136,
  2020{\natexlab{a}}.

\bibitem[Jagtap et~al.(2020{\natexlab{b}})Jagtap, Kawaguchi, and
  Karniadakis]{jagtap2020laaf}
Ameya~D. Jagtap, Kenji Kawaguchi, and George~Em Karniadakis.
\newblock Locally adaptive activation functions with slope recovery for deep
  and physics-informed neural networks.
\newblock \emph{Proceedings of the Royal Society A}, 476\penalty0
  (2239):\penalty0 20200334, 2020{\natexlab{b}}.

\bibitem[Katharopoulos et~al.(2020)Katharopoulos, Vyas, Pappas, and
  Fleuret]{katharopoulos2020linear}
Angelos Katharopoulos, Apoorv Vyas, Nikolaos Pappas, and Fran{\c{c}}ois
  Fleuret.
\newblock Transformers are {RNNs}: Fast autoregressive transformers with linear
  attention.
\newblock In \emph{Proceedings of the 37th International Conference on Machine
  Learning}, volume 119 of \emph{Proceedings of Machine Learning Research},
  pages 5156--5165, 2020.

\bibitem[Khalfaoui-Hassani and Kesselheim(2026)]{khalfaoui2026poly}
Ismail Khalfaoui-Hassani and Stefan Kesselheim.
\newblock Polynomial, trigonometric, and tropical activations.
\newblock In \emph{International Conference on Learning Representations
  (ICLR)}, 2026.
\newblock arXiv:2502.01247.

\bibitem[Klambauer et~al.(2017)Klambauer, Unterthiner, Mayr, and
  Hochreiter]{klambauer2017selu}
G{\"u}nter Klambauer, Thomas Unterthiner, Andreas Mayr, and Sepp Hochreiter.
\newblock Self-normalizing neural networks.
\newblock In \emph{Advances in Neural Information Processing Systems
  (NeurIPS)}, 2017.

\bibitem[Ko et~al.(2026)Ko, Kim, Ham, and Jung]{ko2026tesla}
Daehwa Ko, JaeHyeon Kim, SeungHyun Ham, and Jay~Hoon Jung.
\newblock {TESLA}: Taylor expansion of sinusoidal learnable activations.
\newblock In \emph{Proceedings of the 29th International Conference on
  Artificial Intelligence and Statistics (AISTATS)}, volume 300 of
  \emph{Proceedings of Machine Learning Research}, pages 5257--5265, 2026.

\bibitem[Krizhevsky(2009)]{krizhevsky2009cifar}
Alex Krizhevsky.
\newblock Learning multiple layers of features from tiny images.
\newblock Technical report, University of Toronto, 2009.

\bibitem[LeCun et~al.(1998)LeCun, Bottou, Orr, and
  M{\"u}ller]{lecun1998efficient}
Yann LeCun, L{\'e}on Bottou, Genevieve~B. Orr, and Klaus-Robert M{\"u}ller.
\newblock Efficient {B}ack{P}rop.
\newblock In \emph{Neural Networks: Tricks of the Trade}, volume 1524 of
  \emph{Lecture Notes in Computer Science}, pages 9--50. Springer, 1998.

\bibitem[Liu et~al.(2021)Liu, Lin, Cao, Hu, Wei, Zhang, Lin, and
  Guo]{liu2021swin}
Ze~Liu, Yutong Lin, Yue Cao, Han Hu, Yixuan Wei, Zheng Zhang, Stephen Lin, and
  Baining Guo.
\newblock Swin transformer: Hierarchical vision transformer using shifted
  windows.
\newblock In \emph{IEEE/CVF International Conference on Computer Vision
  (ICCV)}, pages 10012--10022, 2021.

\bibitem[Ma et~al.(2021)Ma, Zhang, Liu, and Sun]{ma2021acon}
Ningning Ma, Xiangyu Zhang, Ming Liu, and Jian Sun.
\newblock Activate or not: Learning customized activation.
\newblock In \emph{IEEE/CVF Conference on Computer Vision and Pattern
  Recognition (CVPR)}, pages 8032--8042, 2021.

\bibitem[Maas et~al.(2013)Maas, Hannun, and Ng]{maas2013leakyrelu}
Andrew~L. Maas, Awni~Y. Hannun, and Andrew~Y. Ng.
\newblock Rectifier nonlinearities improve neural network acoustic models.
\newblock In \emph{ICML Workshop on Deep Learning for Audio, Speech and
  Language Processing}, 2013.

\bibitem[Misra(2020)]{misra2019mish}
Diganta Misra.
\newblock Mish: A self regularized non-monotonic activation function.
\newblock In \emph{British Machine Vision Conference (BMVC)}, 2020.
\newblock arXiv:1908.08681.

\bibitem[Molina et~al.(2020)Molina, Schramowski, and Kersting]{molina2020pau}
Alejandro Molina, Patrick Schramowski, and Kristian Kersting.
\newblock Pad{\'e} activation units: End-to-end learning of flexible activation
  functions in deep networks.
\newblock In \emph{International Conference on Learning Representations
  (ICLR)}, 2020.

\bibitem[Nair and Hinton(2010)]{nair2010relu}
Vinod Nair and Geoffrey~E. Hinton.
\newblock Rectified linear units improve restricted {B}oltzmann machines.
\newblock In \emph{International Conference on Machine Learning (ICML)}, pages
  807--814, 2010.

\bibitem[Pan and Yang(2010)]{pan2010survey}
Sinno~Jialin Pan and Qiang Yang.
\newblock A survey on transfer learning.
\newblock \emph{IEEE Transactions on Knowledge and Data Engineering},
  22\penalty0 (10):\penalty0 1345--1359, 2010.

\bibitem[Ramachandran et~al.(2018)Ramachandran, Zoph, and
  Le]{ramachandran2018swish}
Prajit Ramachandran, Barret Zoph, and Quoc~V. Le.
\newblock Searching for activation functions.
\newblock In \emph{ICLR Workshop}, 2018.
\newblock arXiv:1710.05941.

\bibitem[Scardapane et~al.(2019)Scardapane, Van~Vaerenbergh, Totaro, and
  Uncini]{scardapane2019kaf}
Simone Scardapane, Steven Van~Vaerenbergh, Simone Totaro, and Aurelio Uncini.
\newblock Kafnets: Kernel-based non-parametric activation functions for neural
  networks.
\newblock \emph{Neural Networks}, 110:\penalty0 19--32, 2019.

\bibitem[Shor et~al.(2026)Shor, Litany, and Bronstein]{shor2026spectral}
Tamir Shor, Or~Litany, and Alex~M. Bronstein.
\newblock Learnable spectral activations.
\newblock \emph{arXiv preprint arXiv:2610.07419}, 2026.

\bibitem[Strack et~al.(2024)Strack, Safari, and Hutter]{strack2024search}
Lukas Strack, Mahmoud Safari, and Frank Hutter.
\newblock Efficient search for customized activation functions with gradient
  descent.
\newblock \emph{arXiv preprint arXiv:2408.06820}, 2024.

\bibitem[Tavakoli et~al.(2021)Tavakoli, Agostinelli, and
  Baldi]{tavakoli2021splash}
Mohammadamin Tavakoli, Forest Agostinelli, and Pierre Baldi.
\newblock {SPLASH}: Learnable activation functions for improving accuracy and
  adversarial robustness.
\newblock \emph{Neural Networks}, 140:\penalty0 1--12, 2021.

\bibitem[Teney et~al.(2025)Teney, Jiang, Gogianu, and
  Abbasnejad]{teney2025simplicity}
Damien Teney, Liangze Jiang, Florin Gogianu, and Ehsan Abbasnejad.
\newblock Do we always need the simplicity bias? looking for optimal inductive
  biases in the wild.
\newblock In \emph{Proceedings of the IEEE/CVF Conference on Computer Vision
  and Pattern Recognition (CVPR)}, 2025.

\bibitem[Tolstikhin et~al.(2021)Tolstikhin, Houlsby, Kolesnikov, Beyer, Zhai,
  Unterthiner, Yung, Steiner, Keysers, Uszkoreit, Lucic, and
  Dosovitskiy]{tolstikhin2021mixer}
Ilya Tolstikhin, Neil Houlsby, Alexander Kolesnikov, Lucas Beyer, Xiaohua Zhai,
  Thomas Unterthiner, Jessica Yung, Andreas Steiner, Daniel Keysers, Jakob
  Uszkoreit, Mario Lucic, and Alexey Dosovitskiy.
\newblock {MLP}-mixer: An all-{MLP} architecture for vision.
\newblock In \emph{Advances in Neural Information Processing Systems},
  volume~34, pages 24261--24272, 2021.

\bibitem[Xiao et~al.(2017)Xiao, Rasul, and Vollgraf]{xiao2017fmnist}
Han Xiao, Kashif Rasul, and Roland Vollgraf.
\newblock Fashion-{MNIST}: A novel image dataset for benchmarking machine
  learning algorithms.
\newblock \emph{arXiv:1708.07747}, 2017.

\bibitem[Yosinski et~al.(2014)Yosinski, Clune, Bengio, and
  Lipson]{yosinski2014transferable}
Jason Yosinski, Jeff Clune, Yoshua Bengio, and Hod Lipson.
\newblock How transferable are features in deep neural networks?
\newblock In \emph{Advances in Neural Information Processing Systems
  (NeurIPS)}, pages 3320--3328, 2014.

\end{thebibliography}
